\pdfoutput=1

\documentclass[11pt]{article}

\usepackage[]{acl}

\usepackage{times}
\usepackage{latexsym}
\usepackage{booktabs}
\usepackage{multicol}
\usepackage{multirow}
\usepackage{graphicx}
\usepackage{amsmath}
\usepackage{amssymb}
\usepackage{caption}
\usepackage{subcaption}
\usepackage{makecell}
\usepackage{pifont}
\newcommand{\cmark}{\ding{51}}%
\newcommand{\xmark}{\ding{55}}%
\usepackage{url}
\usepackage[T1]{fontenc}

\usepackage[utf8]{inputenc}

\usepackage{microtype}

\title{Decoder Tuning: Efficient Language Understanding as Decoding}

\author{Ganqu Cui$^1$, Wentao Li$^1$, Ning Ding$^1$, Longtao Huang$^2$, Zhiyuan Liu$^{1,3}$\thanks{\ \ Corresponding Author.}, Maosong Sun$^{1,3}$\footnotemark[1] \\
$^{1}$ NLP Group, DCST, IAI, BNRIST, Tsinghua University, Beijing\\
  $^{2}$ Alibaba Group
  $^{3}$ IICTUS, Shanghai\\
  \texttt{cgq22@mails.tsinghua.edu.cn}}

\begin{document}
\maketitle
\begin{abstract}
With the evergrowing sizes of pre-trained models (PTMs), it has been an emerging practice to only provide the inference APIs for users, namely model-as-a-service (MaaS) setting. To adapt PTMs with model parameters frozen, most current approaches focus on the input side, seeking for powerful prompts to stimulate models for correct answers.
However, we argue that input-side adaptation could be arduous due to the lack of gradient signals and they usually require thousands of API queries, resulting in high computation and time costs.
In light of this, we present Decoder Tuning (DecT), which in contrast optimizes task-specific decoder networks on the output side.
Specifically, DecT first extracts prompt-stimulated output scores for initial predictions. On top of that, we train an additional decoder network on the output representations to incorporate posterior data knowledge. 
By gradient-based optimization, DecT can be trained within several seconds and requires only one PTM query per sample. 
Empirically, we conduct extensive natural language understanding experiments and show that DecT significantly outperforms state-of-the-art algorithms with a $200\times$ speed-up. Our codes are available at \url{https://github.com/thunlp/DecT}.
\end{abstract}

\section{Introduction}

Recent advances in pre-trained models (PTMs) demonstrate the power of the ``pre-training-fine-tuning'' paradigm, which empowers broad downstream NLP tasks with a single backbone model~\cite{Devlin19bert, raffel2020exploring, radford2019language}. Given the million even billion-scale models, model-as-a-service (MaaS) has become an emerging practice in deploying massive PTMs, where users can only get access to model inference APIs~\cite{Brown20language, sun2022bbt}. Under such a scenario, PTMs' parameters are frozen, and users cannot fine-tune the model on downstream tasks for adaptation. To find an alternative way, researchers have studied  MaaS PTM adaptation methods extensively.
\begin{figure}
    \centering
    \includegraphics[width=\linewidth]{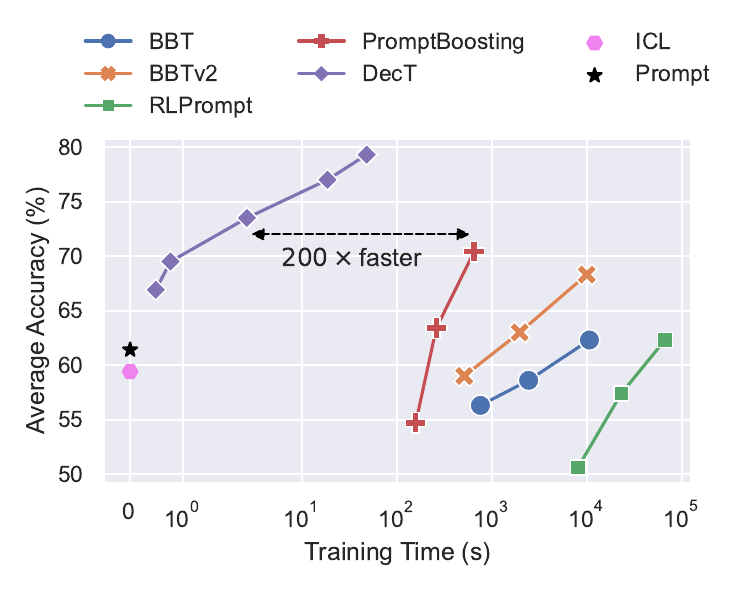}
    \caption{Accuracy v.s. training time for MaaS adaptation methods under different training shots. We plot $\{1, 4, 16, 64, 256\}$-shot DecT, 0-shot Prompt, 1-shot ICL and $\{1, 4, 16\}$-shot for other algorithms. DecT outperforms all baselines by a large margin with a $200\times$speed-up.}
    \label{fig:eff}
\end{figure}

Most existing approaches in this line are based on \textit{prompts}, which modify inputs with specific patterns. By wrapping inputs into cloze-style questions or prepending inputs with a few demonstrative examples, PTMs could produce the right outputs directly and show strong ``in-context'' learning abilities~\cite{Petroni19lama,Brown20language} without any parameter update. 
Besides heuristic prompt design, some recent works try to optimize the input prompts without gradients. Among them, Black-box Tuning (BBT)~\cite{sun2022bbt} and BBTv2~\cite{sun2022bbtv2} apply evolutionary algorithm~\cite{hansen2001completely} on continuous prompt tokens, while RLPrompt~\cite{deng2022rlprompt} adopts reinforcement learning to find discrete prompt tokens. Nevertheless, gradient-free optimization is rather difficult and these input-side methods need to query the PTMs thousands of times for optimization, which leads to huge inference costs in terms of time and computation resources. Moreover, their final performance is not satisfying as well.

Given the flaws of \textit{input-side} adaptation, we turn to \textit{output-side} adaptation, which builds tunable decoder networks on model outputs. Comparatively, output-side adaptation enjoys two major advantages: (1) We can directly tune decoder networks on top of model outputs with back-propagation rather than arduous alternatives. (2) We can reduce thousands of model queries to only once per sample. However, designing decoder networks is not straightforward. Past studies have shown that merely tuning an MLP or LSTM~\cite{hochreiter1997long} over output features cannot provide satisfying results~\cite{sun2022bbtv2, sun2022bbt}, leaving this path underexplored.

In this work, we aim to solve the performance issue for output-side adaptation, and we argue that there are two critical reasons behind it: (1) Simply utilizing PTMs as feature extractors ignores the infilling ability of PTMs, which is a strong prior for adaptation. (2) MLP and LSTM are not proper networks especially when training data is not sufficient. 

Based on these findings, we present Decoder Tuning (DecT), an enhanced output-side adaptation method. Specifically, DecT has two crucial design choices to address the above issues. First,
DecT queries the PTM with prompts and adopts model output scores as the initial predictions, which takes advantage of internal model knowledge. Second, on top of the output representations, we select a Prototypical Network (ProtoNet)~\cite{Snell17proto} as the decoder network and train it to fit the training data, which is more suitable for few-shot learning. In this way, DecT modifies the initial model scores with subsequent training data, thus achieving better performance.


Through few-shot learning experiments on ten language understanding datasets, we highlight three advantages of DecT (see Figure~\ref{fig:eff}). 
(1) DecT achieves over $3\%$ absolute accuracy improvement on average, greatly outperforming previous works. 
(2) DecT is highly efficient. Compared with major prompt engineering baselines, DecT dramatically reduces the average adaptation time from over 9,800 seconds (BBTv2) to 3 seconds.
(3) DecT only requires one PTM query for each example, while other input-side optimization methods need about $10^4$ calls. This advantage is vital when PTM calls are not for free. In addition, we conduct extensive ablation studies and validate the impact of each component of DecT.

\section{Preliminaries}
\label{sec:pre}
Given a set of training data $\mathcal{D}_{\text{train}} = \{ (x_i, y_i)\}_{i=1}^{N}$ and PTM $\mathcal{M}$, we need to predict the label $y\in\{1,\dots, K\}$ for sample $x$, where $K$ is the number of classes. We assume that each class has the same amount of $n$ training samples.

In the MaaS setting, $\mathcal{M}$ is a black-box inference API with fixed parameters. Therefore, we can only query the model with input $x$ and get corresponding outputs. To better utilize the PTMs, it has been a common practice to wrap input samples into prompts. Specifically, we enclose each input $x$ into a template $\mathcal{T}$ with a \texttt{[MASK]} token (here we assume using a masked language model). Then, we query $\mathcal{M}$ with $\mathcal{T}(x)$ and get the final layer hidden states $\mathbf{h}$ at the \texttt{[MASK]} position and scores $\mathbf{s}=S_{\mathcal{M}}(\mathcal{T}(x)) \in \mathbb{R}^{K}$ over label words $\mathcal{V}$. Take sentiment analysis as an example, we can use 
$$
\mathcal{T}(x) = x\ \text{In summary, it was \texttt{[MASK]}.}
$$
as the template with $\mathcal{V} = \{\text{bad}, \text{great}\}$ as label words for negative and positive sentiment respectively. The output scores on these label words further correspond to the classes. 

\section{Methodology}
\begin{figure*}
    \centering
    \includegraphics[width=0.85\linewidth]{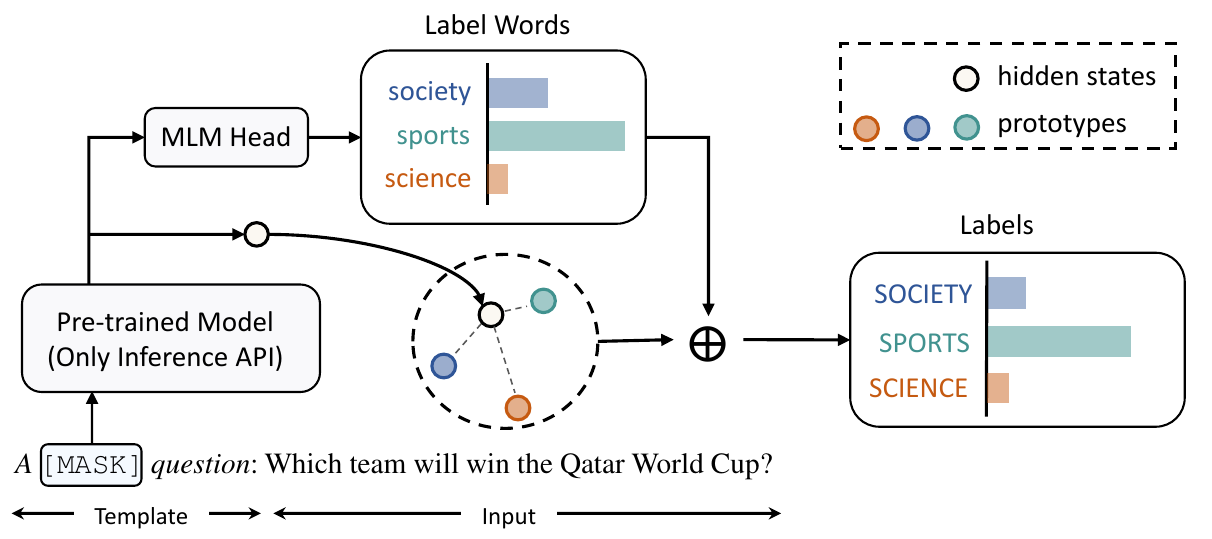}
    \caption{Pipeline of DecT. We feed the PTM with prompts and collect model output scores over a set of label words (\textbf{Top}) and hidden states at \texttt{[MASK]} position. The hidden states are used to train a ProtoNet to fit training data (\textbf{Bottom}). We make final predictions by combining model and ProtoNet scores.}
    \label{fig:model}
\end{figure*}
In this section, we elaborate on our proposed Decoder Tuning (DecT) method for the classification task. We start with reviewing current input-side adaptation methods, then give an overview of DecT and finally detail it step-by-step.

\subsection{Input-side Adaptation}
Previous MaaS adaptation methods seek for optimal prompts that stimulate PTMs to output correct answers\footnote{BBTv2~\cite{sun2022bbtv2} further optimizes prompt tokens in the intermediate layers, but we omit this here.}. Without loss of generality, we formulate these methods with a transformation function $f(\cdot)$ which pre-processes the input $x$. $f(\cdot)$ can be specialized by adding demonstrations~\cite{Brown20language}, discrete prompt tokens~\cite{deng2022rlprompt} or soft ones~\cite{sun2022bbtv2,sun2022bbt}. Denote the final score as $q(x)$ and probability as $P(y|x) = \text{Softmax}(q(x))$, these methods define $q(x)=S_{\mathcal{M}}(f(x))$ and optimize $f(\cdot)$ for correct predictions.
Although optimizing $f(\cdot)$ without model gradients is possible, we argue that it is highly burdensome. Forwarding through a large ``black box'' model $\mathcal{M}$, it is rather challenging to find corresponding inputs for specific outputs without the guidance of gradient signals. As a result, users may get suboptimal performance with expensive query costs. We empirically validate it in experiments. 

\subsection{Overview of DecT}
For more effective and efficient PTM adaptation, we turn to output-side adaptation rather than input-side. 
Overall, output-side adaptation can be viewed as a post-processing of model outputs which uses another function $g(\cdot)$ to process the model outputs, and get the final scores $q(x) = g(S_{\mathcal{M}}(\mathcal{T}(x)))$. Different from input-side ones, output-side adaptation is easy-to-optimize with gradient descent, and for each sample, we only need to query the PTM once.

For DecT, as shown in Figure~\ref{fig:model}, we model the post-processing as decoding, which refers to a post-modification to the initial model predictions. Specifically, we first query the PTM with prompt-enclosed inputs to get model outputs, including the scores for each class and hidden states. Intuitively, output scores contain prior knowledge inside the PTM, so we retain them as part of the final scores.
Then, we tune an additional decoder function on the hidden states to fit the training data and make final predictions. Next, we describe how we query the model and then specify the implementation of the score function. 

\subsection{Querying with Prompts}
To get model outputs, we simply follow the procedure in Section~\ref{sec:pre} and query the model with manual template-wrapped inputs. We then process the scores by calibration.
\paragraph{Calibration.} As stated in \citet{zhao2021calibrate}, PTMs tend to assign higher probabilities on those frequent label words, leading to biased output scores. 
To eliminate the prediction bias, we further calibrate the output scores with empty input $x_{\text{c}}=$``'' following \cite{zhao2021calibrate}. Querying the model with $x_{\text{c}}$, we can obtain the calibaration scores $\mathbf{s}_{c}$ and normalize them by $\mathbf{s}_c/\text{mean}(\mathbf{s}_c)$.
Then we calibrate $\mathbf{s}$ by
\begin{equation}
    \mathbf{\hat{s}} = \text{diag}(\mathbf{s}_c/\text{mean}(\mathbf{s}_c))^{-1}\mathbf{s}.
\end{equation}
After that, the calibrated scores $\mathbf{\hat{s}}$ are balanced over classes.

\subsection{Tuning the Outputs}
After getting the hidden states and calibrated scores, we perform DecT outside the PTM to modify the output scores fitting the training data. Denote the final score on class $k$ as $q(x, k)$, we calculate it by the following function:
\begin{equation}
\label{eq:q}
   q(x, k)  = \text{Dec}(\mathbf{h}, k) + \lambda \mathbf{\hat{s}}_k,
\end{equation}
where $\text{Dec}(\cdot)$ is a trainable decoder function,  $\lambda$ is a hyperparameter controlling the weight of PTM scores and $\mathbf{\hat{s}}_k$ is the $k$-th logit in $\mathbf{\hat{s}}$. By tuning $\text{Dec}(\cdot)$, the final predictions incorporate training data on top of PTM outputs, which combine both knowledge effectively.

The design choice of $\text{Dec}(\cdot)$ is fairly flexible. In practice, we select Prototypical Networks (ProtoNet)~\cite{Snell17proto} due to their simplicity and remarkable performance in few-shot learning and prompt-based tuning~\cite{cui2022prototypical}. For this, we project the hidden states with a linear layer parameterized by $\mathbf{W}$ and get sample representation
\begin{equation}
    \mathbf{v}  = \mathbf{Wh}.
\end{equation}

On prototypes, classical approaches model them as points in the embedding space, which overlook the different class characteristics. Inspired by \citet{ding2022bigp}, we model prototypes as hyperspheres with an additional radius parameter. Concretely, the prototype for class $k$ contains two parameters, center position vector $\mathbf{z}_k$ and radius scalar $r_k$. We randomly initialize $\mathbf{z}_k$ and initialize $r_k$ as the average distance between $\mathbf{z}_k$ and instances in class $k$:
\begin{equation}
    r_k = \frac{1}{N_k}\sum_i^{y_i=k} \Vert\mathbf{v}_i - \mathbf{z}_k\Vert_2.
\end{equation}
As for the score function, we calculate the Euclidean distances between instances and prototypes. 
\begin{equation}
    \text{Dec}(\mathbf{h}, k) = - \Vert\mathbf{Wh}- \mathbf{z}_k\Vert_2 + r_k.
\end{equation}

According to Eq.~\ref{eq:q}, the final logit is
\begin{equation}
    q(x,k)=- \Vert\mathbf{Wh}- \mathbf{z}_k\Vert_2 + r_k +\lambda \mathbf{\hat{s}}_k.
\end{equation}
From a geometric view, the score function calculates the distance from instance $x$ to the ``surface'' of the prototype, where $r_k +\lambda \mathbf{\hat{s}}_k$ is the whole radius acting like the bias term.
With the scores, we can calculate the predicted probability by the Softmax function:
\begin{equation}
    P(y=k|x) = \frac{\exp(q(x, k))}{\sum_{k'=1}^K \exp(q(x, k'))},
\end{equation}
and we can optimize $\mathbf{W}$ and $r_k$ by the cross-entropy loss
\begin{equation}
    \mathcal{L} = -\frac{1}{N}\sum_{i=1}^{N} \log P(y_i|x_i).
\end{equation}
\section{Experiments}
\begin{table*}[]
\resizebox{\textwidth}{!}{%
\begin{tabular}{@{}c|lccccccccccc@{}}
\toprule
\textbf{$n$} &
  \multicolumn{1}{l}{\textbf{Method}} &
  \multicolumn{1}{c}{\textbf{SST2}} &
  \multicolumn{1}{c}{\textbf{IMDB}} &
  \multicolumn{1}{c}{\textbf{Yelp}} &
  \multicolumn{1}{c}{\textbf{AG}} &
  \multicolumn{1}{c}{\textbf{DB}} &
  \multicolumn{1}{c}{\textbf{Yahoo}} &
  \multicolumn{1}{c}{\textbf{RTE}} &
  \multicolumn{1}{c}{\textbf{SNLI}} &
  \multicolumn{1}{c}{\textbf{MNLI-m/mm}} &
  \multicolumn{1}{c}{\textbf{NERD}} &
  \multicolumn{1}{c}{\textbf{Avg.}} \\
\midrule
0 &
  \textbf{Prompt} &
  $83.3_{}$ &
  $89.4_{}$ &
  $87.1_{}$ &
  $80.9_{}$ &
  $68.4_{}$ &
  $49.9_{}$ &
  $52.4_{}$ &
  $40.7_{}$ &
  $50.8_{}/51.7_{}$ &
  $21.1_{}$ &
  $61.4_{}$ \\
\midrule
\multirow{5}{*}{1} &
  \textbf{ICL} &
  $81.5_{3.7}$ &
  $65.6_{11.4}$ &
  $81.1_{10.6}$ &
  $66.7_{4.8}$ &
  $71.7_{2.6}$ &
  $53.2_{6.2}$ &
  $45.0_{4.7}$ &
  $46.1_{5.3}$ &
  ${\bf 53.6_{0.5}/53.9_{0.8}}$ &
  $34.7_{3.3}$ &
  $59.4_{4.9}$ \\
 &
  \textbf{BBT} &
  $83.4_{1.3}$ &
  $89.0_{0.1}$ &
  $89.7_{0.1}$ &
  $75.4_{0.8}$ &
  $59.1_{1.7}$ &
  $31.2_{2.7}$ &
  $52.3_{1.4}$ &
  $38.5_{0.8}$ &
  $43.4_{2.5}/42.9_{3.3}$ &
  $14.1_{2.3}$ &
  $56.3_{1.5}$ \\
 &
  \textbf{BBTv2} &
  $83.3_{2.5}$ &
  $89.0_{0.2}$ &
  $89.9_{0.2}$ &
  $74.3_{3.2}$ &
  $74.2_{5.2}$ &
  $34.0_{3.5}$ &
  $48.2_{5.7}$ &
  $38.6_{4.0}$ &
  $44.2_{3.2}/44.3_{4.5}$ &
  $29.0_{0.8}$ &
  $59.0_{3.0}$ \\
 &
  \textbf{RLPrompt} &
  $63.5_{6.3}$ &
  $65.0_{6.5}$ &
  $66.3_{6.9}$ &
  $72.5_{4.5}$ &
  $65.6_{5.5}$ &
  $38.1_{5.8}$ &
  $53.8_{5.3}$ &
  $36.5_{3.0}$ &
  $40.3_{2.0}/41.0_{2.1}$ &
  $14.5_{1.8}$ &
  $50.6_{4.7}$ \\
 &
 \textbf{PromptBoosting} &
  $86.7_{2.6}$ &
  $82.4_{6.1}$ &
  $88.7_{2.5}$ &
  $58.7_{11.8}$ &
  $73.0_{4.8}$ &
  $23.7_{7.0}$ &
  $50.0_{5.9}$ &
  $43.5_{6.1}$ &
  $36.8_{1.6}/36.3_{2.3}$ &
  $22.0_{0.8}$ &
  $54.7_{4.6}$ \\
 &
  \textbf{DecT} &
  ${\bf 90.8_{0.2}}$ &
  ${\bf 91.2_{0.3}}$ &
  ${\bf 94.8_{0.1}}$ &
  ${\bf 79.9_{1.1}}$ &
  ${\bf 78.8_{0.9}}$ &
  ${\bf 55.2_{0.8}}$ &
  ${\bf 56.0_{2.7}}$ &
  ${\bf 47.7_{4.1}}$ &
  $52.2_{2.7}/53.3_{3.0}$ &
  ${\bf 35.7_{1.5}}$ &
  ${\bf 66.9_{1.6}}$ \\
\midrule
\multirow{5}{*}{4} &
  \textbf{ICL} &
  $60.3_{9.8}$ &
  $80.4_{6.6}$ &
  $77.4_{14.6}$ &
  $65.1_{5.4}$ &
  $71.7_{6.5}$ &
  $49.9_{9.9}$ &
  $42.7_{3.9}$ &
  $42.1_{3.2}$ &
  $44.7_{5.9}/45.2_{6.0}$ &
  $31.7_{4.8}$ &
  $55.6_{7.0}$ \\
 &
  \textbf{BBT} &
  $84.5_{1.2}$ &
  ${\bf 89.8_{0.9}}$ &
  $90.2_{0.6}$ &
  $79.0_{2.1}$ &
  $67.7_{3.5}$ &
  $42.9_{0.6}$ &
  $48.4_{4.0}$ &
  $40.5_{1.3}$ &
  $41.2_{1.7}/40.7_{2.0}$ &
  $19.4_{1.5}$ &
  $58.6_{1.8}$ \\
 &
  \textbf{BBTv2} &
  $86.6_{2.2}$ &
  $89.4_{0.6}$ &
  $90.3_{0.5}$ &
  $79.1_{2.1}$ &
  $89.0_{1.7}$ &
  $46.0_{1.4}$ &
  $46.2_{2.3}$ &
  $40.8_{4.3}$ &
  $44.0_{0.9}/44.8_{1.6}$ &
  $31.9_{1.4}$ &
  $62.6_{1.7}$ \\
 &
  \textbf{RLPrompt} &
  $80.7_{7.5}$ &
  $75.8_{10.1}$ &
  $78.8_{7.3}$ &
  $76.1_{4.8}$ &
  $76.3_{5.9}$ &
  $45.0_{3.1}$ &
  $53.5_{2.9}$ &
  $36.3_{2.6}$ &
  $44.4_{2.9}/45.5_{3.8}$ &
  $16.7_{2.4}$ &
  $57.4_{4.8}$ \\
 &
 \textbf{PromptBoosting} &
  ${\bf88.9_{2.3}}$ &
  $83.0_{5.2}$ &
  $92.3_{2.1}$ &
  $78.2_{6.8}$ &
  ${\bf90.1_{0.7}}$ &
  $36.4_{5.1}$ &
  $53.5_{5.9}$ &
  ${\bf53.4_{3.4}}$ &
  $39.8_{4.5}/40.3_{5.7}$ &
  $40.9_{2.5}$ &
  $63.4_{4.0}$ \\
 &
  \textbf{DecT} &
  $87.6_{1.6}$ &
  $89.6_{0.9}$ &
  ${\bf 94.8_{0.7}}$ &
  ${\bf 81.9_{2.6}}$ &
  $89.1_{0.6}$ &
  ${\bf 59.9_{2.1}}$ &
  ${\bf 56.7_{2.7}}$ &
  $53.2_{2.9}$ &
  ${\bf 52.2_{2.3}/53.4_{2.4}}$ &
  ${\bf 46.7_{1.7}}$ &
  ${\bf 69.5_{1.9}}$ \\
\midrule
\multirow{5}{*}{16} &
  \textbf{ICL} &
  $71.5_{15.8}$ &
  $80.6_{6.0}$ &
  $73.7_{14.5}$ &
  $64.4_{6.0}$ &
  $71.8_{9.1}$ &
  $52.6_{5.7}$ &
  $43.8_{7.0}$ &
  $42.0_{6.3}$ &
  $51.4_{3.0}/52.1_{3.3}$ &
  $35.1_{2.6}$ &
  $58.1_{7.2}$ \\
 &
  \textbf{BBT} &
  $89.6_{0.3}^b$ &
  $89.3_{0.4}$ &
  $91.5_{0.2}^b$ &
  $81.5_{0.8}^b$ &
  $87.8_{3.0}^b$ &
  $48.3_{1.4}$ &
  $52.6_{2.2}^b$ &
  $46.6_{1.3}^b$ &
  $40.0_{2.6}/39.9_{2.9}$ &
  $17.8_{1.4}$ &
  $62.3_{1.5}$ \\
 &
  \textbf{BBTv2} &
  $90.3_{1.7}^a$ &
  $88.6_{2.1}$ &
  $92.9_{0.6}^a$ &
  $85.3_{0.5}^a$ &
  $93.6_{0.7}^a$ &
  $52.0_{1.4}$ &
  $56.7_{3.3}^a$ &
  $57.3_{2.3}^a$ &
  $50.1_{2.4}/51.7_{3.2}$ &
  $33.3_{1.0}$ &
  $68.3_{1.7}$ \\
 &
  \textbf{RLPrompt} &
  $87.0_{2.6}$ &
  $87.6_{2.4}$ &
  $95.1_{1.0}^c$ &
  $80.2_{0.7}^c$ &
  $80.8_{3.3}$ &
  $48.1_{2.2}$ &
  $54.3_{2.8}$ &
  $41.1_{5.0}$ &
  $43.3_{3.9}/44.3_{4.5}$ &
  $17.5_{1.4}$ &
  $61.8_{2.7}$ \\
 &
  \textbf{PromptBoosting} &
  $87.6_{3.0}^d$ &
  $86.2_{3.1}$ &
  $94.7_{1.0}$ &
  $85.2_{0.9}^d$ &
  ${\bf95.0_{0.5}}$ &
  $46.6_{2.4}$ &
  ${\bf60.0_{5.5}^d}$ &
  ${\bf61.3_{3.5}^d}$ &
  $52.5_{1.5}^d/50.4_{5.1}$ &
  $52.1_{2.6}$ &
  $70.1_{2.6}$ \\
 &
  \textbf{DecT} &
  ${\bf91.0_{0.5}}$ &
  ${\bf 91.0_{0.9}}$ &
  ${\bf 95.4_{0.3}}$ &
  ${\bf 86.4_{0.4}}$ &
  $94.6_{0.5}$ &
  ${\bf 64.2_{0.7}}$ &
  $59.7_{1.8}$ &
  $60.5_{0.8}$ &
  ${\bf 55.3_{1.3}/56.8_{1.5}}$ &
  ${\bf 53.5_{1.8}}$ &
  ${\bf 73.5_{1.0}}$ \\
\bottomrule

\end{tabular}%
}
\caption{Experiment results for MaaS adaptation methods. Some baseline results are taken from corresponding papers ($^a$\citet{sun2022bbtv2}, $^b$\citet{sun2022bbt}, $^c$\citet{deng2022rlprompt},
$^d$\citet{hou2022promptboosting}). We run other experiments over 5 random seeds and report average accuracy and standard deviation (\%). Best results are in \textbf{bold}.
}
\label{tab:results}
\end{table*}
In this section, we first introduce the experimental settings (Section~\ref{sec:exp-setting}), then discuss the results for few-shot experiments (Section~\ref{sec:exp-results}), efficiency comparison (Section~\ref{sec:eff}), and experiment results for more training data (Section~\ref{sec:ft}).

\subsection{Experimental Settings}
\label{sec:exp-setting}
\paragraph{Datasets.} We conduct experiments on four typical natural language understanding tasks. For sentiment analysis, we select SST2~\cite{socher2013recursive}, Yelp P.~\cite{zhang2015character} and IMDB~\cite{maas2011learning}. For text classification, we use AG's News, Yahoo~\cite{zhang2015character} and DBPedia~\cite{lehmann2015dbpedia}. For natural language inference (NLI), we adopt RTE~\cite{dagan2005pascal, haim2006second, giampiccolo2007third, bentivogli2009fifth}, SNLI~\cite{bowman2015large} and MNLI~\cite{williams2018broad}. For entity typing, we experiment on FewNERD~\cite{ding21fewnerd}. We report dataset statistics in Appendix~\ref{sec:data}.
\paragraph{Splits.} We randomly sample $n=1,4,16$ data instances for each class from the training set for few-shot learning, and sample same amount data for validation. For datasets in GLUE~\cite{wang2019glue} (SST2, RTE, MNLI) and SNLI, we use the original validation sets as test sets following \citet{zhang2021revisiting}. For other datasets, we evaluate on their original test sets.
\paragraph{Baselines.} We compare with representative MaaS PTM adaptation methods. \textbf{Prompt} refers to directly performing zero-shot classification with template-wrapped examples. \textbf{In-context learning (ICL)}~\cite{Brown20language} further concatenates some exemplars before the test samples. \textbf{BBT}~\cite{sun2022bbt} optimizes soft prompt tokens with an evolutionary algorithm, and \textbf{BBTv2}~\cite{sun2022bbtv2} further inserts deep prompts to intermediate layers for better performance.
\textbf{RLPrompt}~\cite{deng2022rlprompt} is another recent algorithm that optimizes discrete prompts with reinforcement learning.
\textbf{PromptBoosting}~\cite{hou2022promptboosting} is a concurrent work that applies boosting algorithm for prompt ensembling.
We report the details of baselines in Appendix~\ref{sec:baseline}.
\paragraph{Environments.} For all experiments, we use NVIDIA A100 and RTX 2080 Ti GPUs. We implement DecT with PyTorch~\cite{paszke19pytorch}, HuggingFace Tansformers~\cite{wolf2020transformers}, and OpenPrompt~\cite{ding2021openprompt}.
\paragraph{Implementation Details.} For all methods, we use the same RoBERTa$_{\texttt{LARGE}}$~\cite{liu2019roberta} as the backbone model. For DecT, we set the representation dimension to 128 and optimize the parameters for 30 epochs with Adam optimizer~\cite{Kingma14adam}. The learning rate is 0.01. On the selection of $\lambda$, we directly set $\lambda=1/n$ for most datasets based on the intuition that $\lambda$ should decrease as the amount of training data increases. On MNLI and FewNERD, we tune $\lambda$ on the validation set and select $\lambda=1$ and $\lambda=1/16$ respectively. We give the templates and label words in Appendix~\ref{app:templates}.

\subsection{Main Results}
\label{sec:exp-results}

Table~\ref{tab:results} presents the main few-shot learning results. From the results, we have these observations: 

\textbf{Overall, DecT outperforms the state-of-the-art baseline methods by a large margin (more than $3\%$ on average),} especially under extreme data scarcity, showing its superior performance. Across different tasks, DecT and baselines obtain similar results on some easy sentiment analysis and topic classification tasks, but we highlight that DecT is much more favorable on difficult datasets, such as Yahoo and FewNERD. While other baseline methods struggle to optimize well, DecT surpasses them significantly (about $10\%$ on Yahoo and $20\%$ on FewNERD under 16-shot setting compared with BBTv2 and ICL). 

On stability, DecT also has consistently low variance and some baselines (ICL, RLPrompt and PromptBoosting) are  unstable. Given the difficulty of few-shot PTM adaptation, it is of great significance that the adaptation method is robust to random seeds.

On baselines, optimization-free methods, i.e. zero-shot prompt and ICL are strong baselines. However, as shown in the table, ICL gives the best results in the 1-shot setting, and it can hardly improve with more training data due to the input length restriction.
To compare, merely optimizing the input prompts (BBT and RLPrompt) can hardly outperform them, showing the limitation of input-side prompt optimization. In contrast, two other baselines, BBTv2 and PromptBoosting, are more powerful because they either 
inserts additional learnable prompt tokens inside the PTM or ensembles the outputs of different prompts.
With the superior results of DecT, we argue that output-side optimization is a promising way for MaaS PTM adaptation.

\subsection{Efficiency Comparison}
\label{sec:eff}
\begin{table}[]
\resizebox{\linewidth}{!}{%
\begin{tabular}{@{}lccc@{}}
\toprule
\textbf{Method}   & \textbf{Tr. Time (s)} & \textbf{\# Query} & \textbf{\# Param. (K)} \\ \midrule
\textbf{ICL}      & 0                          & 0                 & 0                     \\
\textbf{BBT}      & 10,512                     & 8,000             & 0.5                   \\
\textbf{BBTv2}    & 9,856                     & 8,000             & 12                      \\
\textbf{RLPrompt} & 65,579                     & 12,000            & 3,100                   \\
\textbf{PromptBoosting} & 644 & 10 & 0.4 \\
\textbf{DecT}     & 3                          & 1                 & 130                     \\ \bottomrule
\end{tabular}%
}
\caption{Efficiency comparison of MaaS adaptation methods. Training time is the average wall clock time measured in the 16-shot setting.
``Tr.'' stands for Training and ``Param.'' stands for Parameter.}
\label{tab:eff}
\end{table}

\begin{table*}[t]
\resizebox{\textwidth}{!}{%
\begin{tabular}{@{}c|lccccccccccc@{}}
\toprule
\textbf{$n$} &
  \multicolumn{1}{l}{\textbf{Method}} &
  \multicolumn{1}{c}{\textbf{SST2}} &
  \multicolumn{1}{c}{\textbf{IMDB}} &
  \multicolumn{1}{c}{\textbf{Yelp}} &
  \multicolumn{1}{c}{\textbf{AG}} &
  \multicolumn{1}{c}{\textbf{DB}} &
  \multicolumn{1}{c}{\textbf{Yahoo}} &
  \multicolumn{1}{c}{\textbf{RTE}} &
  \multicolumn{1}{c}{\textbf{SNLI}} &
  \multicolumn{1}{c}{\textbf{MNLI-m/mm}} &
  \multicolumn{1}{c}{\textbf{NERD}} &
  \multicolumn{1}{c}{\textbf{Avg.}} \\
\midrule
\multirow{2}{*}{64} &
  \textbf{Fine-tuning$^\dagger$} &
  $92.5_{1.9}$ &
  $86.3_{3.8}$ &
  $94.5_{1.4}$ &
  $87.4_{0.6}$ &
  $98.2_{0.2}$ &
  $69.0_{0.7}$ &
  $67.7_{3.2}$ &
  $66.6_{6.4}$ &
  $65.6_{2.9}/67.7_{4.0}$ &
  $67.6_{0.8}$ &
  $78.5_{2.4}$ \\
 &
  \textbf{DecT} &
  $92.4_{0.5}$ &
  $91.3_{0.5}$ &
  $94.9_{0.5}$ &
  $89.2_{0.3}$ &
  $97.0_{0.1}$ &
  $69.3_{0.4}$ &
  $65.7_{1.7}$ &
  $67.2_{1.0}$ &
  $62.0_{1.4}/63.3_{1.3}$ &
  $56.1_{0.8}$ &
  $77.1_{0.8}$ \\
\midrule
\multirow{2}{*}{256} &
  \textbf{Fine-tuning$^\dagger$} &
  $92.0_{0.9}$ &
  $92.1_{0.2}$ &
  $94.3_{0.3}$ &
  $89.6_{0.3}$ &
  $98.5_{0.2}$ &
  $70.2_{0.4}$ &
  $79.8_{1.0}$ &
  $84.4_{0.4}$ &
  $77.2_{0.2}/78.7_{0.3}$ &
  $71.4_{0.5}$ &
  $84.4_{0.4}$ \\
 &
  \textbf{DecT} &
  $92.7_{0.2}$ &
  $92.1_{0.1}$ &
  $95.6_{0.1}$ &
  $90.3_{0.1}$ &
  $97.4_{0.1}$ &
  $71.3_{0.1}$ &
  $69.2_{1.0}$ &
  $69.7_{0.4}$ &
  $68.0_{0.3}/69.4_{0.3}$ &
  $56.2_{0.3}$ &
  $79.3_{0.3}$ \\

\bottomrule

\end{tabular}%
}
\caption{Experiment results for more training data.  We run all experiments over 5 random seeds and report the average accuracy and standard deviation (\%). $^\dagger$: Update model parameters.
}
\label{tab:ft}
\end{table*}

Despite the superior performance, another major advantage of DecT is its high efficiency. In Figure~\ref{fig:eff}, we plot average accuracy versus training time for each method under different shots. We also provide detailed statistics of training time, query numbers, and parameter numbers for 16-shot experiments in Table~\ref{tab:eff}.

From Figure~\ref{fig:eff} and Table~\ref{tab:eff}, we clearly see that DecT can be optimized quickly and only requires one model query per training sample, \textbf{which is about $200\times$faster and queries $10\times$fewer than all prompt optimization methods}. For BBT, BBTv2, and RLPrompt, users have to query the model near $10^4$ times and spend several hours for sufficient optimization even in the few-shot scenario. When the inference API is not for free such as OpenAI API~\footnote{\url{https://openai.com/api/}}, using these methods would be expensive, and this further burdens their usage in the scenarios of rich data and large models.

In terms of tunable parameters, DecT demands 130K additional parameters for the linear projection layer, which is less than $0.04\%$ of RoBERTa$_{\texttt{LARGE}}$ (355M) that largely saves storage space. Although some other methods (BBT, BBTv2 and PromptBoosting) require fewer parameters, DecT is much easier to optimize.

\subsection{Beyond Few-shot}
\label{sec:ft}
As shown in Section~\ref{sec:eff}, the simple architecture and high efficiency enable DecT to scale on more training data, while baseline methods struggle to finish training within acceptable time limits. To explore the scalability of DecT beyond the few-shot setting, we conduct experiments with increased training data ($n=64$ and $256$). For reference, we compare DecT with fine-tuning, the strongest baseline which update full model parameters. 

The detailed results are presented in Figure~\ref{fig:eff} and Table~\ref{tab:ft} and we have the following conclusions. (1) DecT continually improves its performance on more training data at a low cost. The average accuracy gains $6\%$ from 16-shot to 256-shot while the average training time is less than $100$ seconds. 
(2) Compared with fine-tuning, DecT is even on par with it in the 64-shot scenario and gradually falls behind in the 256-shot setting, which is reasonable as we only tune a small portion of parameters outside the model. Through further task-level observation, we find DecT still performs well on sentiment analysis and topic classification, but cannot catch up with fine-tuning on NLI and entity typing, which are identified as harder tasks as they require complex reasoning or fine-grained semantic understanding.
(3) In experiments, we find fine-tuning is more sensitive to random seeds in the few-shot setting due to the huge amount of trainable parameters and relatively few loss signals, which is evidenced by the high variance in the 64-shot setting. In such scenario, DecT has lower variances due to most parameters are frozen. Therefore, the stability advantage of DecT has been verified again.

To conclude, we take the first step to applying MaaS methods beyond few-shot learning. \textbf{The results show that DecT is competitive against fine-tuning on regular classification tasks, but is limited on difficult tasks.} How to adapt PTMs on challenging tasks without parameter updates still needs further exploration.

\section{Analysis}
In addition to main experiments, we further provide more analytical experiments for understanding DecT. We conduct ablation study on several components in Section~\ref{sec:ablation}. Then we evaluate the scaling effect (Section~\ref{sec:ptm}), the impact of hyperparameter $\lambda$ (Section~\ref{sec:lambda}) and templates (Section~\ref{sec:template}) respectively. We further conduct transferability experiments in Appendix~\ref{sec:transfer}.

\subsection{Ablation Study}
\label{sec:ablation}
\begin{table}[]
    \centering
    \begin{tabular}{cc|ccc}
    \toprule
       \multirow{2}{*}{$\mathbf{s}$} & \multirow{2}{*}{$r$} & \multicolumn{3}{c}{\textbf{Average Accuracy}} \\
    \cmidrule{3-5}
      & & 1 & 4 & 16 \\
    \midrule
       \xmark & \xmark & $54.0_{6.3}$ & $66.3_{2.5}$ & $73.0_{1.2}$ \\
       \cmark & \xmark & $64.8_{2.6}$ & $69.3_{1.7}$ & ${\bf 73.5_{1.0}}$\\
       \xmark & \cmark & $54.0_{6.2}$ & $67.2_{2.0}$ & $73.0_{1.1}$\\
       \cmark & \cmark & ${\bf 66.9_{1.6}}$ & ${\bf 69.5_{1.9}}$ & ${\bf 73.5_{1.0}}$ \\
    \bottomrule
    \end{tabular}
    \caption{Ablation study of model scores $\mathbf{s}$ and radius parameter $r$. We run each experiment over 5 random seeds and report average accuracy and standard deviation (\%). Best results are in \textbf{bold}.}
    \label{tab:ablation}
\end{table}
\begin{figure}
    \centering
    \includegraphics[width=.8\linewidth]{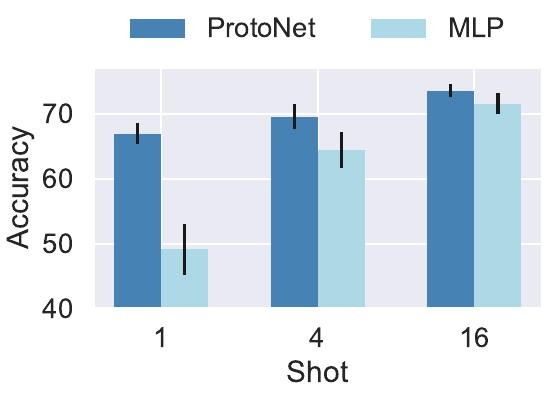}
    \caption{Comparison between ProtoNet and MLP. We report the average accuracy (\%) and standard deviation.}
    \label{fig:mlp}
\end{figure}
To validate each component of our proposed DecT, especially the effect of model scores $\mathbf{s}$, radius parameter $r$, and ProtoNet, we conduct extensive ablation studies. We present results in Table~\ref{tab:ablation} and Figure~\ref{fig:lambda}.

\paragraph{Ablating model scores.} Apparently, model scores contribute largely to the few-shot performance of DecT, especially when the training data is extremely scarce (1-shot), which illustrates that model scores contain beneficial prior model knowledge for language understanding. Also, incorporating training data reduces the variance. When there are more training data, model scores bring less enhancement, which is reasonable as the relative weights of model and ProtoNet scores change.
\paragraph{Ablating radius.} Meanwhile, the radius is also helpful under low-shot scenarios, which characterizes the difference across classes. But as the number of training data increases, ProtoNet dominates model predictions and the impact of $r$ is weakened as well.
\paragraph{Ablating decoder.} As stated previously, the design choice of the decoder function is flexible. We replace ProtoNet with a two-layer MLP and evaluate the performance. In Figure~\ref{fig:mlp} we can see that ProtoNet significantly outperforms MLP in the 1-shot setting, which matches the advantages of ProtoNet in the few-shot setting. In 4-shot and 16-shot experiments, ProtoNet still gets higher scores, but with smaller margins. On stability, ProtoNet achieves consistently lower standard deviation scores, which serve as another advantage. Overall, we find ProtoNet is a vital component in DecT, and simply replacing it with MLP would worsen the performance.

\begin{table*}[h]
\resizebox{\textwidth}{!}{%
\begin{tabular}{@{}llllllllllll@{}}
\toprule
\textbf{Model} &
  \multicolumn{1}{c}{\textbf{SST2}} &
  \multicolumn{1}{c}{\textbf{IMDB}} &
  \multicolumn{1}{c}{\textbf{Yelp}} &
  \multicolumn{1}{c}{\textbf{AG}} &
  \multicolumn{1}{c}{\textbf{DB}} &
  \multicolumn{1}{c}{\textbf{Yahoo}} &
  \multicolumn{1}{c}{\textbf{RTE}} &
  \multicolumn{1}{c}{\textbf{SNLI}} &
  \multicolumn{1}{c}{\textbf{MNLI-m/mm}} &
  \multicolumn{1}{c}{\textbf{NERD}} &
  \multicolumn{1}{c}{\textbf{Avg.}} \\
\midrule
\textbf{T5$_{\texttt{Small}}$} &
  $73.4_{1.8}$ &
  $68.8_{2.1}$ &
  $79.5_{1.4}$ &
  $79.1_{0.6}$ &
  $76.8_{0.7}$ &
  $57.5_{0.7}$ &
  $51.9_{0.8}$ &
  $38.7_{2.1}$ &
  $38.6_{0.4}/39.0_{0.3}$ &
  $35.1_{1.9}$ &
  $58.0_{1.2}$ \\
\textbf{T5$_{\texttt{Base}}$} &
  $83.8_{1.1}$ &
  $86.3_{0.9}$ &
  $91.5_{0.6}$ &
  $84.3_{0.6}$ &
  $93.5_{0.4}$ &
  $61.1_{0.8}$ &
  $54.0_{1.6}$ &
  $44.9_{1.6}$ &
  $47.8_{0.6}/49.4_{0.7}$ &
  $50.2_{3.0}$ &
  $67.9_{1.1}$ \\
\textbf{T5$_{\texttt{Large}}$} &
  $92.3_{0.4}$ &
  $92.0_{0.4}$ &
  $94.4_{1.3}$ &
  $85.5_{0.8}$ &
  $94.9_{0.3}$ &
  $63.6_{0.9}$ &
  $55.9_{2.3}$ &
  $49.5_{1.9}$ &
  $49.7_{1.4}/50.8_{1.9}$ &
  $53.2_{1.6}$ &
  $71.1_{1.2}$ \\
\textbf{T5$_{\texttt{3B}}$} &
  $89.9_{0.4}$ &
  $92.7_{0.7}$ &
  $94.9_{2.0}$ &
  $87.7_{0.8}$ &
  $96.2_{0.3}$ &
  $66.5_{0.7}$ &
  $55.8_{2.2}$ &
  $52.0_{1.9}$ &
  $52.8_{1.6}/52.2_{2.1}$ &
  $51.9_{1.4}$ &
  $72.1_{1.3}$ \\
\bottomrule
\end{tabular}%
}
\caption{Experiment results (16-shot) for our method on different versions of T5~\cite{raffel2020exploring}. We run each experiment over 5 random seeds and report average accuracy and standard deviation (\%).
}
\label{tab:scaling}
\end{table*}

\begin{table*}[h]
\resizebox{\textwidth}{!}{%
\begin{tabular}{@{}llllllllllll@{}}
\toprule
\textbf{Shot} &
  \multicolumn{1}{c}{\textbf{SST2}} &
  \multicolumn{1}{c}{\textbf{IMDB}} &
  \multicolumn{1}{c}{\textbf{Yelp}} &
  \multicolumn{1}{c}{\textbf{AG}} &
  \multicolumn{1}{c}{\textbf{DB}} &
  \multicolumn{1}{c}{\textbf{Yahoo}} &
  \multicolumn{1}{c}{\textbf{RTE}} &
  \multicolumn{1}{c}{\textbf{SNLI}} &
  \multicolumn{1}{c}{\textbf{MNLI-m/mm}} &
  \multicolumn{1}{c}{\textbf{NERD}} &
  \multicolumn{1}{c}{\textbf{Avg.}} \\
\midrule
\textbf{0} &
  $80.5$ &
  $89.1$ &
  $96.6$ &
  $74.6$ &
  $71.3$ &
  $46.7$ &
  $84.1$ &
  $45.4$ &
  $45.6/45.6$ &
  $1.6$ &
  $61.9$ \\
\textbf{4} &
  $91.2$ &
  $96.5$ &
  $97.8$ &
  $80.5$ &
  $81.1$ &
  $56.5$ &
  $82.2$ &
  $77.8$ &
  $66.0/66.5$ &
  $52.9$ &
  $77.2$ \\
\textbf{16} &
  $92.7$ &
  $96.2$ &
  $97.5$ &
  $85.5$ &
  $89.8$ &
  $65.2$ &
  $86.0$ &
  $86.4$ &
  $76.3/76.3$ &
  $54.6$ &
  $82.4$ \\
\textbf{64} &
  $94.3$ &
  $96.5$ &
  $98.3$ &
  $88.5$ &
  $93.5$ &
  $68.7$ &
  $87.1$ &
  $88.9$ &
  $78.0/79.0$ &
  $59.8$ &
  $84.8$ \\
\textbf{256} &
  $94.5$ &
  $96.7$ &
  $98.4$ &
  $89.7$ &
  $94.2$ &
  $69.9$ &
  $87.7$ &
  $89.4$ &
  $81.7/80.6$ &
  $59.1$ &
  $85.6$ \\
\bottomrule
\end{tabular}%
}
\caption{Experiment results for our method on CPM-Bee. We run each experiment over 5 random seeds and report average accuracy (\%).
}
\label{tab:cpmbee}
\end{table*}

\subsection{Model Scaling}
\label{sec:ptm}
In this section, we explore how DecT applies to PTMs with different architecture and scales.\\
We select T5~\cite{raffel2020exploring}, an encoder-decoder PTM, at different scales, from T5$_{\texttt{Small}}$, T5$_{\texttt{Base}}$, T5$_{\texttt{Large}}$ to T5$_{\texttt{3B}}$. 
Table~\ref{tab:scaling} presents the full results of T5 models. First of all, DecT has been successfully deployed on T5, a generative language model, which verifies its transferability across PTMs. Furthermore, we can observe an apparent trend of the scaling effect where larger models consistently perform better.\\
We also evaluate the DecT on CPM-Bee\footnote{\url{https://live.openbmb.org/models/bee}}, which is a bilingual generative pre-trained language model with 10B parameters. Table~\ref{tab:cpmbee} presents the results of CPM-Bee in different settings. The results show that DecT strongly enhances the adaptation of large PLM on downstream tasks. Moreover, CPM-Bee achieves great performance on NLI tasks, which flags that DecT could deal with more difficult tasks with powerful backbone models.

\subsection{Impact of $\lambda$}
\label{sec:lambda}
\begin{figure}
    \centering
    \begin{subfigure}[b]{.49\linewidth}
         \centering
         \includegraphics[width=\textwidth]{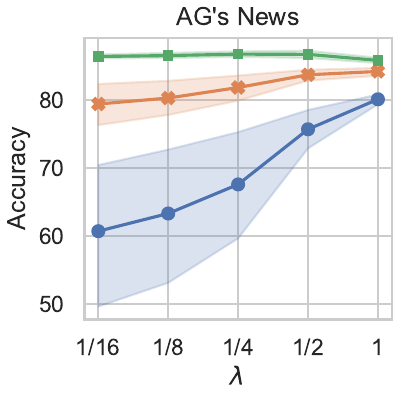}
     \end{subfigure}
     \begin{subfigure}[b]{.49\linewidth}
         \centering
         \includegraphics[width=\textwidth]{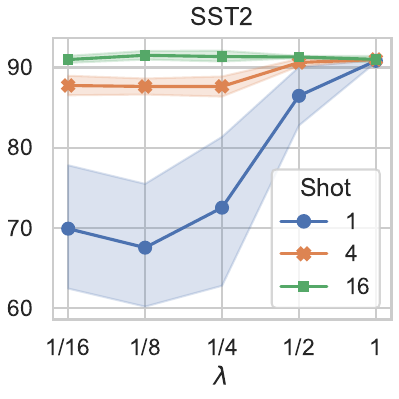}
     \end{subfigure}
     \caption{Accuracy (\%) with 85\% confidence interval on 5 runs for different $\lambda$ on AG's News and SST2.}
     \label{fig:lambda}
\end{figure}

As a hyperparameter, $\lambda$ controls the relative importance of model scores and prototype scores. Here we examine its impact on AG's News and SST2. In Figure~\ref{fig:lambda}, we can observe that: (1) $\lambda$ largely affects DecT in the 1-shot settings. As $\lambda$ increases, DecT gradually performs better and stabler, which illustrates the importance of model knowledge in this case. (2) With the shot number increases, the impact of varying $\lambda$ is weakened, and the best practices become smaller. These observations validate our selection strategy in Section~\ref{sec:exp-setting}, which effectively balances model and data knowledge. 

\subsection{Impact of Templates}
\label{sec:template}
Although DecT is an output-side adaptation method, the choice of templates also affects the final performance. To assess the influence of templates, we conduct experiments on AG's News and SST2 and show results in Table~\ref{tab:exp-temp}. Overall, DecT does not rely much on templates. While different templates may induce fluctuant zero-shot performance, DecT largely moderates the gaps between them. Additionally, we try two templates searched from RLPrompt~\cite{deng2022rlprompt} and they both achieve satisfying results. On SST2, the template from RLPrompt is even better than manually designed ones. \textbf{Therefore, we highlight that DecT is complementary with input-side adaptation algorithms, and they can work together for better performance.}

\begin{table}[]
\resizebox{\linewidth}{!}{%
\begin{tabular}{@{}lcc@{}}
\toprule
\textbf{Template}                                                      & \multicolumn{1}{c}{\textbf{Prompt}} & \multicolumn{1}{c}{\textbf{DecT}} \\
\midrule
\multicolumn{3}{c}{SST2}\\
\midrule
$x$ In summary, it was \texttt{[MASK]}.        & 83.3                       & 91.0                     \\
$x$ It was \texttt{[MASK]}.  & 73.3                       & 88.4                     \\
\makecell[l]{$x$ AgentMediaGrade\\Officials Grade \texttt{[MASK]}.$^c$} & 90.4                       & 92.2                     \\
\midrule
\multicolumn{3}{c}{AG's News}\\
\midrule
{[} Topic : \texttt{[MASK]} {]} $x$    & 80.9                       & 86.4                     \\
{[} Category : \texttt{[MASK]} {]} $x$ & 78.6                       & 86.8                     \\
\makecell[l]{\texttt{[MASK]} Alert Blog Dialogue\\ Diary Accountability $x$ $^c$} & 78.8                       & 86.0                   \\
\bottomrule
\end{tabular}
}
\caption{Accuracy of prompt (zero-shot) and DecT (16-shot) across different templates. Templates marked with $^c$ are taken from \citet{deng2022rlprompt}.}
\label{tab:exp-temp}
\end{table}
\section{Related Work}
Our work explores how to efficiently adapt large PTMs. In this section, we review three lines of research for \textit{prompt-based tuning} (data efficiency), \textit{parameter-efficient tuning} (parameter efficiency), and MaaS adaptation methods respectively.

\subsection{Prompt-based Tuning}
Prompt-based tuning aims to bridge the gap between pre-training and downstream tasks for data-efficient model adaptation.
The major practice for prompt-based tuning~\cite{liu2021pre} is wrapping text pieces into human-designed templates. By this means, prompt-based tuning converts downstream tasks to pre-training tasks (e.g. masked language modeling) and greatly enhances the zero/few-shot learning ability of PTMs. Prompt-based tuning is first applied in knowledge probing~\cite{trinh2018simple,Petroni19lama}, and it has been adopted broadly in NLP~\cite{schick21pet, hu2021knowledgeable, ding2021prompt, han2021ptr, cui2022prototypical}. Despite manually designed prompts, other works also investigate automatic and learnable prompts~\cite{shin2020autoprompt, gao20making, hambardzumyan21warp, schick20petal, lester2021power} to alleviate the prompt engineering efforts. 
However, the optimization of prompts is a non-trivial problem~\cite{ding2022delta, lester2021power} which sometimes leads to more computation costs and suboptimal performance. Thus in our work, we adopt manual prompts to stimulate model knowledge and help data-efficient model adaptation.

\subsection{Parameter-efficient Tuning}
Another line of work explores tuning a small fraction of model parameters to reduce computation and storage budgets, namely parameter-efficient tuning (PET)~\cite{ding2022delta}. Typical PET methods include inserting tunable modules~\cite{Houlsby19, LiL2021prefix, hu2022lora}, adding soft prompt tokens~\cite{ lester2021power} and specifying certain parameters~\cite{Zaken22bitfit}. Although PET methods achieve remarkable performance with few parameter updates, they still require model gradients, which are unavailable in the MaaS setting.

\subsection{MaaS Adaptation}
With inference-only APIs, there are also works that adapt models without tuning any model parameters. \citet{Brown20language} present in-context learning, which concatenates test inputs with several exemplars. Further improvements focus on reliving the instability issues caused by model biases~\cite{zhao2021calibrate, han2022prototypical} and examplar orders~\cite{lu2022fantastically}. PromptBoosting~\cite{hou2022promptboosting} employs boosting algorithm for prompt ensembling, giving strong performance.
Other approaches try to optimize prompts with either black-box optimization methods~\cite{sun2022bbtv2, sun2022bbt} or reinforcement learning~\cite{deng2022rlprompt}. However, due to the lack of gradient signals, they need thousands of model queries, resulting in high costs when the model is large and API calls are not for free. Different from the abovementioned methods, we adapt models at the output side, which need not optimize the distant input prompts. We demand only one API call for each training sample and achieve better results across tasks. 
\section{Conclusion}
In this paper, we present DecT, which performs both data and parameter-efficient adaptation with off-shelf PTMs. By fusing prior model knowledge and posterior data knowledge, DecT achieves superior performance on ten language understanding tasks. Meanwhile, DecT exceeds existing baselines by three orders of magnitude in terms of training time and the number of queries, highlighting its advantages in real-world deployment.
In future works, we are eager to explore how to combine input and output-side adaptation methods for better PTM adaptation, and how to extend this line of research to more challenging scenarios.

\section*{Limitation}
DecT explores how to adapt black-box PTMs on downstream tasks. As we show in Section~\ref{sec:ft}, our method is not comparable to fine-tuning on hard tasks with increased data points. Moreover, we only focus on classification tasks in this work and do not testify DecT on free-form generation tasks. In the future, we will work toward more general MaaS adaptation strategies across tasks.

\section*{Ethical Statement}
As large language models are getting more and more popular in NLP research and application, DecT provides a cost-efficient way to adapt these large models. However,  we need also to be cautious about the improper adaptation of large language models, such as generating toxic and biased speeches. 

\section*{Acknowledgements}
This work is supported by the National Key R\&D Program of China (No.2022ZD0116312)  and Institute Guo Qiang at Tsinghua University.

Ganqu made the original research proposal and wrote the paper. 
Ganqu Cui and Wentao Li conducted experiments.
Ning Ding revised the paper and participated in the discussion. Longtao Huang, Zhiyuan Liu and Maosong Sun advised the project.

\bibliography{custom}
\bibliographystyle{acl_natbib}
\clearpage
\appendix
\section{Experiment Details}
\subsection{Dataset Statistics}
\label{sec:data}
We provide dataset statistics in Table~\ref{tab:dataset}. We obtain AG's News, Yahoo, and Yelp P. from \url{https://github.com/zhangxiangxiao/Crepe} under the BSD-3-Clause license. We get FewNERD from \url{https://ningding97.github.io/fewnerd/} with CC BY-SA 4.0 license. Other datasets are downloaded using Huggingface Datasets~\cite{lhoest-etal-2021-datasets}.

\begin{table}[h]
\resizebox{\linewidth}{!}{%
\begin{tabular}{ccccc}
\toprule
\textbf{Task}                                  & \textbf{Dataset}        & \textbf{\# Class}  & \textbf{\# Test}   \\
\midrule
\multirow{3}{*}{\makecell[c]{Sentiment\\ Analysis}}   & SST2         & 2                  & 872       \\
                                      & Yelp         & 2                  & 38,000     \\
                                      & IMDB           & 2                  & 25,000     \\ \midrule
\multirow{3}{*}{\makecell[c]{Topic\\ Classification}} & AG's News         & 4                  & 7,600      \\
                                      & Yahoo          & 10                 & 60,000     \\
                                      & DBPedia        & 14                 & 70,000     \\ \midrule
\multirow{3}{*}{NLI}                  & RTE            & 2                  & 277       \\
                                      & SNLI           & 3                  & 9,842      \\
                                      & MNLI-m/mm & 3                  & 9,815/9,832 \\ \midrule
Entity Typing                         & FewNERD      & 66                 & 96,853    \\ 
\bottomrule
\end{tabular}
}
\caption{The dataset information of our experiments. \# Class column represents the number of classes, \# Test column represents the number of examples for testing.
}
\label{tab:dataset}
\end{table}

\subsection{Baselines}
\label{sec:baseline}
\paragraph{In-context Learning (ICL).} To guarantee meaningful results, we randomly permute the demonstrations and prepend them before input prompts. Due to the input length limit, we truncate the demonstrations which exceed input length. 
\paragraph{BBT and BBTv2.} We reproduce BBT and BBTv2 using the official codes\footnote{\url{https://github.com/txsun1997/Black-Box-Tuning}}. For datasets adopted in their work, we follow the original implementations including templates, label words, and hyperparameters. For other datasets, we reproduce with our templates, label words, and default hyperparameters. We take existing 16-shot experiment results from the paper and run other experiments with 5 random seeds for a fair comparison.
\paragraph{RLPrompt.} We also use the official codes\footnote{\url{https://github.com/mingkaid/rl-prompt}} for reproduction and take some experiment results from their original paper. It is worth noting that RLPrompt adopts the test set of SST2 and we use the validation set, so we report the reproduced results on the SST2 validation set.
\paragraph{PromptBoosting.} We follow the official codes\footnote{\url{https://github.com/UCSB-NLP-Chang/PromptBoosting}} for reproduction. Since the number of additional parameters is related to number of classes, we compute the average numbers across datasets.
\paragraph{Fine-tuning.} We adopt prompt-based fine-tuning which uses the same templates and label words with DecT. We tune the whole model for 5 epochs with AdamW optimizer~\cite{loshchilov19decoupled} using a $3\times 10^{-5}$ learning rate.
\subsection{Templates and Label Words}
\label{app:templates}
We report the used prompt templates and label words in Table~\ref{tab:template}. Most of them are taken from OpenPrompt~\cite{ding2021openprompt}.
\begin{table*}[h]
\resizebox{\linewidth}{!}{%
\begin{tabular}{lll}
\toprule
\textbf{Dataset}        & \textbf{Template}                                          & \textbf{Label Words}\\ \midrule
SST2          & \multirow{3}{*}{$x$ In summary, it was \texttt{[MASK]}.} & \multirow{3}{*}{bad, great}
\\
Yelp         &                                                   &
\\ 
IMDB           &                                                   &
\\ \hline
AG's News         & {[} Topic : \texttt{[MASK]} {]} $x$     & \makecell[l]{ politics, sports, \\ business, technology}
\\  \hline
Yahoo          &                 {[} Topic : \texttt{[MASK]} {]} $x$                                  & \makecell[l]{society, science, health, education,\\ computers, sports, business,\\ entertainment, family, politics}                                                     \\ \hline
DBPedia     &  

 $x_1$ $x_2$ The category of $x_1$  is \texttt{[MASK]}.     & \makecell[l]{company, school, artist, athlete,\\ politics, transportation, building,\\ river, village, animal, plant,\\ album, film, book}
 \\ \hline
RTE            & \multirow{3}{*}{$x_1$ ? \texttt{[MASK]}, $x_2$}              & No, Yes                    \\  
SNLI           &                                                   & No, Maybe, Yes             \\ 
MNLI-m/mm &                                                   &  No, Maybe, Yes                      \\ \hline
FewNERD      & $x$ \texttt{[ENT]} is \texttt{[MASK]}.          & \makecell[l]{actor, director, artist, athlete, politician,\\ scholar, soldier, person, show, religion,\\ company, team,  school, government, media,\\ party, sports, organization, geopolitical, road,\\ water, park, mountain, island, location,\\ software, food, game, ship, train, plane, car,\\ weapon, product, theater, facility, airport,\\ hospital, library, hotel, restaurant, building, \\ championships, attack, disaster, election, \\protest, event, music, written, film, \\painting, broadcast, art, biology, chemical,\\ living, astronomy, god, law, award, disease,\\ medical, language, currency, educational} \\
\bottomrule
\end{tabular}
}
\caption{The templates and label words used in our experiments. For each dataset, $x$, $x_1$, and $x_2$ represent the input sentences or sentence pairs.
\texttt{[ENT]} copies the entity mentioned in the input sentence.}
\label{tab:template}
\end{table*}

\section{Transferability}
\label{sec:transfer}
\begin{figure}
    \centering
    \includegraphics[width=.8\linewidth]{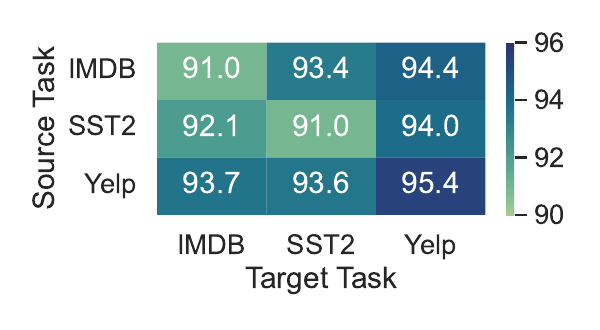}
    \caption{Accuracy heatmap of DecT transferred across different sentiment analysis tasks under 16-shot setting. The backbone model is RoBERTa$_\texttt{LARGE}$}
    \label{fig:transfer}
\end{figure}
We conduct transferability experiments on sentiment analysis and present the results in Figure~\ref{fig:transfer}. We see that DecT is highly transferable across datasets. On SST2 and IMDB, DecT trained on other datasets even surpasses the original performance. More surprisingly, we find DecT trained on Yelp, a restaurant review dataset, performs best on SST2 and IMDB, which are two movie review datasets. This result further shows the great domain generalization ability of DecT.

\end{document}